\UseRawInputEncoding
\documentclass{ifacconf}

\usepackage{natbib}        

\usepackage[dvipdfmx]{graphicx}
\usepackage{graphics} 
\usepackage[fleqn]{amsmath} 
\usepackage{amssymb}  
\usepackage{bm}
\usepackage{algorithm}
\usepackage{algpseudocode}
\usepackage{comment}
\usepackage{color}
\usepackage{url}

\usepackage{extarrows}
\usepackage{mathrsfs}
\usepackage{latexsym}
\def\qed{\hfill $\Box$}

\newtheorem{assumption}{Assumption}
\newtheorem{lemma}{Lemma}

\newcommand{\dd}{v}
\newcommand{\ww}{w}

\newcommand{\CC}[2]{\mathsf{CC}(#1,#2)}
\newcommand{\CCp}[2]{\mathsf{CC}^{\prime}(#1,#2)}

\newcommand{\XX}{\mathbb{R}^n}
\newcommand{\etaspade}{q}
\newcommand{\etaclub}{p}

\newcommand{\ie}{\emph{i.e.}}

\algnewcommand{\Initialize}[1]{%
	\State \textbf{Initialize:}
	\State \hspace*{\algorithmicindent}\parbox[t]{0.8\linewidth}{\raggedright #1}
}

\begin{document}
\begin{frontmatter}

\title{Automatic Exploration Process Adjustment for Safe Reinforcement Learning \\with Joint Chance Constraint Satisfaction\thanksref{0}}


\author[1]{Yoshihiro Okawa} 
\author[1]{Tomotake Sasaki} 
\author[2]{Hidenao Iwane\thanksref{footnoteinfo}}

\address[1]{Artificial Intelligence Laboratory, FUJITSU LABORATORIES LTD., 
	4-1-1 Kamikodanaka, Nakahara-ku, Kawasaki, Kanagawa 211-8588, Japan (e-mail: okawa.y@fujitsu.com, tomotake.sasaki@fujitsu.com)}
\address[2]{Research $\&$ IT Division, Reading Skill Test, Inc., 2-1-4 Shinkawa, Chuo-ku, Tokyo 104-0033, Japan (email: iwane@rstest.co.jp)}

\thanks[0]{\copyright~2020 the authors. {\it This work has been accepted to IFAC for publication under a Creative Commons Licence CC-BY-NC-ND.}}
\thanks[footnoteinfo]{This work is done while he worked at Fujitsu Laboratories Ltd.}

\begin{abstract}                
In reinforcement learning (RL) algorithms, exploratory control inputs are used during learning to acquire knowledge for decision making and control, while the true dynamics of a controlled object is unknown. However, this exploring property sometimes causes undesired situations by violating constraints regarding the state of the controlled object. In this paper, we propose an automatic exploration process adjustment method for safe RL in continuous state and action spaces utilizing a linear nominal model of the controlled object. Specifically, our proposed method automatically selects whether the exploratory input is used or not at each time depending on the state and its predicted value as well as adjusts the variance-covariance matrix used in the Gaussian policy for exploration. We also show that our exploration process adjustment method theoretically guarantees the satisfaction of the constraints with the pre-specified probability, that is, the satisfaction of a joint chance constraint at every time. Finally, we illustrate the validity and the effectiveness of our method through numerical simulation.
\end{abstract}


\end{frontmatter}
\section{Introduction} \label{Intro}
Reinforcement learning (RL) methods are used to acquire knowledge for decision making and control, \ie, ``policy" or ``control law" in online data-driven manners. In order to learn appropriate policies with RL methods, it is necessary to use exploratory control inputs, and they sometimes result in undesired situations. This is not a serious issue in the problems in which situations can be reset or recovered easily such as the video games \citep{Mnih+15a} or strategy board games \citep{Silver+18a}; however, we cannot ignore this in many kinds of engineering problems. For addressing this issue, RL methods guaranteeing the ``safety" during learning are demanded. 

RL methods guaranteeing the ``safety" have been discussed in literatures with different definitions of safety and some of them are called ``safe reinforcement learning (safe RL) methods". According to the survey paper by \cite{Garcia+15a}, approaches for safe RL are classified into two fundamental categories; ``transforming the optimization criterion" and ``modifying the exploration process". This paper adopts the latter one. In particular, we deal with the safety based on the constraints which are explicitly defined in control problems. More specifically, we define the probability of satisfaction of the constraints regarding the state of the controlled object as the quantitative index to evaluate the safety. In this paper, we propose an automatic exploration process adjustment method for a Gaussian policy guaranteeing that the probability noted above is equal to or greater than its pre-specified lower bound, which is a kind of chance constraint \citep{Shapiro+14a}. Our method automatically selects whether the exploratory input is used or not at each time depending on the state and its predicted value as well as adjusts the variance-covariance matrix of a normal distribution utilizing a linear nominal model of the controlled object and upper bounds of its approximation error. We show that an RL algorithm combined with our exploration process adjustment method theoretically guarantees the satisfaction of the joint chance constraint with the pre-specified probability.

The rest of this paper is organized as follows. In Section~\ref{Prob}, we describe a problem formulation of this paper. Subsequently, in Section~\ref{Main}, we introduce our automatic exploration process adjustment method for the Gaussian policy and show a theoretical guarantee of probabilistic satisfaction of the constraints. In this section, we also show one concrete example of safe RL algorithms with our exploration process adjustment method. In addition, we compare  our study with some related work for safe RL. In Section~\ref{Simu}, we verify the validity and effectiveness of our method through numerical simulations, and finally, we conclude this paper in Section~\ref{Conc}.

\section{Problem Formulation} \label{Prob}
Consider a discrete-time affine nonlinear system  given by
\begin{align}
	x_{k+1}=f(x_k) +g(x_k)u_k,
	\label{eq:true_model}
\end{align}
where $x_k\in\mathbb{R}^{n}$ and $u_k\in\mathbb{R}^{m}$ are the state and the control input at time $k$, respectively, and $f: \mathbb{R}^{n} \to \mathbb{R}^{n}$ and $g: \mathbb{R}^{n}  \to \mathbb{R}^{n\times m}$ are unknown nonlinear functions. We assume that the state $x_k$ can be observed directly, and also, an instantaneous cost $c_{k}\in\mathbb{R}$ given by
\begin{align}
	c_{k+1} = \ell(x_{k},u_{k})
	\label{eq:cost_func}
\end{align}
as well. Here $\ell: \mathbb{R}^n \times \mathbb{R}^m \rightarrow [0, \infty)$ is an unknown instantaneous cost function. As described in the left hand side of (\ref{eq:cost_func}), we denote the instantaneous cost by $c_{k+1}$ following the convention adopted in \cite{Sutton+18a}, while it is produced with the state $x_k$ and the input $u_k$ at time $k$.

The objective of control is to minimize the cumulative discounted cost $J$ given by 
\begin{align}
	J=\sum_{k=0}^{\infty} \gamma^{k} c_{k+1},
	\label{eq:cost}
\end{align}
where $\gamma \in (0,1]$ is a discount rate.

In addition, we consider the situation where desirable conditions about the state is given by the following linear inequalities: 
\begin{align}
	H x\preceq \dd,
	\label{eq:const}
\end{align}
where $\dd=[\dd_1,\ldots,\dd_{n_c}]^\mathrm{T} \in\mathbb{R}^{n_c}$ and $H=\left[h_1,\ldots,h_{n_c}\right]^\mathrm{T} \in\mathbb{R}^{n_c \times n}$, and the symbol $\preceq$ represents that every inequality $\le$ on $\mathbb{R}$ is satisfied regarding each component in the vector. We suppose that the inequalities (\ref{eq:const}) is known, and we define the safety in this paper based on them. 

Let the set $\mathcal{X}\subset\mathbb{R}^n$ be
\begin{align}
	\mathcal{X}:=\left\{x \in \mathbb{R}^n \,|\, H x\preceq \dd\right\}, 
\end{align}
and $\mathcal{X}^\mathrm{int}$ be the interior point set of $\mathcal{X}$. We suppose the existence of the state $x^{*} \in \mathcal{X}^\mathrm{int}$ which satisfies $f(x^{*}) = x^{*}$ and $\ell(x^{*},0) = 0$, while we do not suppose that $x^{*}$ is known. This means that some kind of consistency is held between the cost and safety in the control problem, and the state satisfying both of them is maintainable. For simplicity, we also suppose that the initial state $x_0$ satisfies $x_{0} \in \mathcal{X}$.

Next, we describe the goal of this paper. The common goal of RL methods is to acquire a policy, which corresponds to a control law in control engineering, to minimize (or maximize) evaluation functions related to the instantaneous cost (or the instantaneous reward) obtained at each time. If we know both complete dynamics of the controlled object and the evaluation function, the problem leads to the usual optimal control problem. However, in many cases the above-mentioned information is not completely known beforehand, and the usual optimal control methods cannot attain desired performance. A fundamental appeal of RL methods is that  we can obtain the optimal policy or at least an improved one even in such a situation by updating the current policy with the information obtained online.

In order to acquire an appropriate policy while the properties of the controlled object are unknown, we have to select exploratory inputs during learning which are not optimal according to the current policy. This means that, the possibility of violating desirable conditions given by  (\ref{eq:const}) increases because of the exploration.

In general, stochastic policies are used in RL methods to carry out the above exploration, and thus, the state $x_{k}$ of the controlled object also transits stochastically. To evaluate the satisfaction of the conditions given by (\ref{eq:const}) quantitatively, we consider the (joint) chance constraint
\begin{align}
	& \mathrm{Pr}\left\{ H x_{k} \preceq \dd \right\} \geq \eta.
	\label{eq:cc} \\
	(\Leftrightarrow \ & \mathrm{Pr}\left\{ h_{j}^\mathrm{T} x_{k} \preceq \dd_{j}, \forall j=1,2,\cdots,n_c \right\} \geq \eta) \nonumber
\end{align}
In the above inequality, $\mathrm{Pr}\{\cdot\}$ denotes the probability of the satisfaction of the inequalities in $\{\cdot\}$. We call this ``probability of constraint satisfaction" in the rest of this paper,  and use it as the metric of safety. For simplicity, we assume $\eta\in (0.5, 1)$.

The goal of this paper is to propose an automatic exploration process adjustment method for safe RL which guarantees the satisfaction of the joint chance constraint (\ref{eq:cc}) at every time $k \ge 1$ with $\eta$ determined before learning.\footnote{In this paper, we describe the chance constraint (\ref{eq:cc}) as a joint chance constraint, while it is not a ``joint" one if $n_c=1$.}

Throughout this paper, we assume that the following two most basic conditions are satisfied.
\begin{assumption}\normalfont \label{assumption1}
	The following linear approximate (nominal) model of the nonlinear system in (\ref{eq:true_model}) is known:
	\begin{align}
		x_{k+1} \simeq A x_k +B u_k, 
		\label{eq:lin}
	\end{align}
	where $A\in\mathbb{R}^{n \times n}$ and $B\in\mathbb{R}^{n \times m}$.
\end{assumption}
\vspace{1ex}

Let $e(\cdot, \cdot; f,g,A,B):\mathbb{R}^{n} \times \mathbb{R}^{m} \rightarrow \mathbb{R}^{n}$ be the following function which denotes the approximation error between the nonlinear system given in (\ref{eq:true_model}) and its linear approximate model given in (\ref{eq:lin}):
\begin{align}
	&e(x,u;f,g,A,B)
	:=f(x)+g(x)u-\left(Ax+Bu\right)\\
	&=:
	[e_1(x,u;f,g,A,B),\dots,e_n(x,u;f,g,A,B)]^\mathrm{T}.
	\label{eq:err_fun}
\end{align}
We make the next assumption regarding this function.
\begin{assumption}\normalfont \label{assumption2}
	Regarding the functions $e_i(\cdot,\cdot; f,g,A,B)$, $i = 1,2,\dots,n$ defined in (\ref{eq:err_fun}), scalar values $\bar{e}_i\in[0,\infty)$, $i=1,2,\ldots,n$ satisfying the following relation are known:
	\begin{align}
		\bar{e}_i \ge \sup_{x \in \XX,u \in \mathbb{R}^{m}}\left| e_i(x,u;f,g,A,B) \right|. \label{dfn_e_bar}
	\end{align}
\end{assumption}
An example satisfying this assumption is given in Section~\ref{subsec:simcon}.

Under Assumption~\ref{assumption2}, let $\mathcal{E} \subset \mathbb{R}^{n}$ be the set of the vectors  $\epsilon = [\epsilon_{1},\dots, \epsilon_{n}]^\mathrm{T} \in \mathbb{R}^{n}$ whose elements are
\begin{align}
	\epsilon_{i} = \bar{e}_{i} \textrm{ or } - \bar{e}_{i}.
\end{align}
Note that the size of the set $\mathcal{E}$ is $2^{n}$ by definition.

Assumptions~\ref{assumption1} and \ref{assumption2} mean that we do not know the exact system dynamics but have certain level of prior knowledge about it  somehow (e.g. by physical considerations). We introduce three other assumptions right before we use them in the next section. 

We describe our proposed method in detail and show a theorem which theoretically guarantees the joint chance constraint satisfaction in the following section. 
\begin{rem}\normalfont
	In order to simplify the notations in the rest of this paper, we generalize the joint chance constraint \eqref{eq:cc} and denote $\mathrm{Pr}\left\{ H \xi \preceq \dd \right\} \geq \lambda$ for an arbitrary $\xi \in \mathbb{R}^{n}$ and $\lambda \in (0.5, 1)$ by $\CC{\xi}{\lambda}$. That is, 
	\begin{align}
		\CC{\xi}{\lambda} \xLeftrightarrow{\textrm{def}} \mathrm{Pr}\left\{ H \xi \preceq \dd \right\} \geq \lambda.
		\label{eq:CC}
	\end{align}
	For example, with this notation, the joint chance constraint (\ref{eq:cc}) can be denoted by $\CC{x_{k}}{\eta}$. 
	
	In addition, it is sometimes difficult to deal with the joint chance constraint (\ref{eq:cc}) directly since it is a condition with respect to the probability of the simultaneous satisfaction of all $n_c$ constraints. Therefore, we show some supplemental theoretical results with respect to individual constraints before showing our main result in Theorem~\ref{theorem1}. For this purpose, similar to the notation $\CC{\cdot}{\cdot}$,  we denote $\left\{\mathrm{Pr}\left\{ h_j^\mathrm{T}\xi \leq \dd_j \right\} \geq \lambda, \forall j=1,\ldots,n_c\right\}$ for an arbitrary $\xi \in \mathbb{R}^{n}$ and $\lambda \in (0.5,1)$ by $\CCp{\xi}{\lambda}$. That is,
	\mathindent=.5em
	\begin{align}
		\CCp{\xi}{\lambda} \xLeftrightarrow{\textrm{def}} \left\{\mathrm{Pr}\left\{ h_j^\mathrm{T}\xi \leq \dd_j \right\}\!\geq\!\lambda, \forall j=1,\ldots,n_c\right\}.
		\label{eq:CCp}
	\end{align}
	\mathindent=2em
	Regarding the relationship between the above two kinds of chance constraints (\ref{eq:CC}) and (\ref{eq:CCp}), we can easily prove based on Bonferroni's inequality that $\CCp{\xi}{1-\frac{1-\lambda}{n_c}}$ is a sufficient condition for $\CC{\xi}{\lambda}$.
\end{rem}

\section{Main Result}\label{Main}
As described in Section~\ref{Intro}, approaches for safe RL are classified into two fundamental categories; ``transforming the optimization criterion" and ``modifying the exploration process". We adopt the latter one. In the rest of this section, Subsection~\ref{subsec_no_exp} shows the case in which the exploration is completely removed in the RL algorithm as preliminaries. Next, based on the results shown in Subsection~\ref{subsec_no_exp}, we introduce an automated exploration process adjustment method for the Gaussian policy and show a theorem which gives our method a theoretical guarantee regarding the satisfaction of the joint chance constraint in Subsection~\ref{sec:gausp}. Finally, in Subsection~\ref{SafeRLAlg}, we show an example of safe RL methods based on our exploration process adjustment method. We compare our study with related work for safe RL in Section~\ref{Rela}.

\subsection{Chance constraint satisfaction regarding the input without exploration}\label{subsec_no_exp}

When we consider satisfaction of the constraints by modifying the exploration process in an RL algorithm, it is natural to use an input without exploration in some situations. However, even if we do so, the satisfaction of the chance constraint is not always guaranteed. Therefore, we have to select the input carefully even in such a situation. In this subsection, we show two particular cases in which the input can be selected to guarantee the satisfaction of the chance constraint.

Firstly, let us consider the case in which the function $f$ in (\ref{eq:true_model}) obeys the following assumption. 
\begin{assumption}\normalfont \label{assumption0}
	If $x\in\mathcal{X}$ holds, then $f(x)\in \mathcal{X}$ holds．
\end{assumption}
\vskip 1ex
It is straightforward to see $\mathrm{Pr} \{ H x_{k+1} \!\preceq\! \dd \} \!=\! 1$ holds if we use $u_{k} = 0$ for $x_{k} \in \mathcal{X}$ in this case.

Next, we consider the case in which the particular inputs to return the state into $\mathcal{X}$ from the outside of $\mathcal{X}$ can be obtained. Specifically, we make the following assumption.
\begin{assumption}\normalfont \label{assum_tau}
	Suppose that $x_k = x \notin\mathcal{X}$ at time $k\geq 1$.
	Regardless of time $k$ and the state $x$, 
	we can make the state be in $\mathcal{X}$ within $\tau$ step
	with the particular successive inputs. 
	That is, we previously know the successive inputs $u_{k}^{back},u_{k+1}^{back},\dots,u_{k+j-1}^{back}$ or its calculation procedure 
	to let the state be $x_{k + j} \in\mathcal{X}$ $(1\leq j \le  \tau)$ for any $k$ and $x$.
\end{assumption}
\vskip 1ex
In this case, we obtain the following lemma.
\begin{lemma} \label{lem:tau} 
	Let Assumption~\ref{assum_tau} hold. In addition, we assume that $\CC{x_{k+1}}{\etaclub}$ is satisfied if $x_k \in \mathcal{X}$. Then, $\CC{x_{k+1}}{\etaclub^\tau}$ is satisfied if $x_0 \in \mathcal{X}$.
\end{lemma}
\begin{pf}
	At time $k$, we define ``$x_k \in \mathcal{X}$" as State $1$ and ``$x_{k-i} \in \mathcal{X}\land (x_{k-i+1}, \ldots, x_k \not\in \mathcal{X})$" as State $i+1$.	With this definition, the state transition is expressed by 	the Markov chain in Proposition~\ref{prop:tau} described in Appendix~\ref{sec:proof_tau}, and thus,	this lemma is proven with $x_0 \in \mathcal{X}$ and $\rho_1 \geq \etaclub$.\qed
\end{pf}

\subsection{Automatic exploration process adjustment method for Gaussian policy} \label{sec:gausp}

Next, we propose an automatic exploration process adjustment method which includes usage of the inputs with exploration. In this method, we basically generate the inputs based on a Gaussian policy, which can be applied to problems in continuous state and action space. Specifically, we use the following Gaussian probability density function as the policy function: 
\begin{align}
	&\Pi(u \, | \, x ; \; \ww, \Sigma ) \nonumber \\
	&=\! \frac{1}{(\sqrt{2\pi})^m\sqrt{|\Sigma|}}
	\exp\!\left(\!-\frac{1}{2}(u\!-\!\mu(x;\ww))^\mathrm{T}\Sigma^{-1}\bullet\right), 
	\label{eq:policy}
\end{align}
where $\bullet$ represents the omission of $(u-\mu(x;\ww))$ and $\ww\in \mathbb{R}^{N_{\ww}}$ is the policy parameter.  We generate an input $u$ stochastically according to the $m$-dimensional normal distribution with the mean $\mu(x;\ww) \in \mathbb{R}^m$ and the variance-covariance matrix $\Sigma \in \mathbb{R}^{m \times m}$. We express this as
\begin{align}
	u\sim\mathcal{N}\left(\mu(x;\ww),\Sigma\right).
	\label{eq:u}
\end{align}
The degree of exploration, in other words, how different input is selected from its mean,  depends on $\Sigma$.

Let $\mu_{k}$ denote the mean of the input corresponding to the state $x_k$ of the controlled object at time $k$ and the policy parameter  $\ww_k$, that is, 
\begin{align}
	\mu_k:=\mu(x_k;\ww_k).
	\label{eq:mu_k}
\end{align}
Now we restrict the variance-covariance matrix $\Sigma$ to the following diagonal matrix determined by $\sigma^2$ $(\sigma>0)$:
\begin{align}
	\Sigma
	= \sigma^{2} I_{m}, 
	\label{eq:Sigma}
\end{align}
where $I_{m}$ is an $m \times m$ identity matrix.

Consider the case in which the following assumption holds between the linear nominal model of the controlled object and the constraints.
\begin{assumption}\normalfont \label{assumption4}
	The coefficient matrix $B$ of the nominal model in (\ref{eq:lin}) 
	and $H=\left[h_1,\ldots,h_{n_c}\right]^\mathrm{T}$ in the constraints in (\ref{eq:const}) satisfies
	\begin{align}
		h_{j}^\mathrm{T} B \neq 0,\ \  \forall j = 1,2,\dots, n_{c}.
	\end{align}
\end{assumption}
\vskip 2ex

In this case, we have the following lemma regarding the relationship between the standard deviation $\sigma$ of the Gaussian policy and the satisfaction of the joint chance constraint.
\begin{lemma} \label{cor6} 
	Let Assumptions~\ref{assumption1}, \ref{assumption2} and \ref{assumption4} hold, and let the input $u_k$ be selected according to the normal distribution with mean $\mu_k$ and variance-covariance matrix  $\sigma^{2} I_{m}$. In addition, assume $ h_{j}^\mathrm{T} (A x_{k} + B \mu_{k} + \epsilon)  < \dd_{j}, \forall j = 1,2,\dots,n_{c}$. Then, 
	{\mathindent=10pt
		\begin{align}
			\sigma =\min_{j, \epsilon}
			\frac{1}{\|h_{j}^\mathrm{T}B\|_2 \Phi^{-1}({\etaspade})} 
			\left\{ \dd_j\!-\!h_{j}^\mathrm{T}(Ax_{k}\!+\!B\mu_k \!+\!\epsilon) \right\}
	\end{align}}
	is a sufficient condition for $\CCp{x_{k+1}}{\etaspade}$ being held with $q\in(0.5,1)$. Here $\Phi(\cdot)$ is the cumulative distribution function of a zero mean unit variance Gaussian random variable.
\end{lemma}

\begin{pf}
	The conclusion follows from Corollary~\ref{cor5} in Appendix~\ref{proof_prop} and the assumption $ h_{j}^\mathrm{T} (A x_{k} + B \mu_{k} + \epsilon)  < \dd_{j}$, $\forall j = 1,2,\dots,n_{c}$.\qed
\end{pf}

\begin{figure}[t]
	\centering	
	\includegraphics[clip,width=8cm]{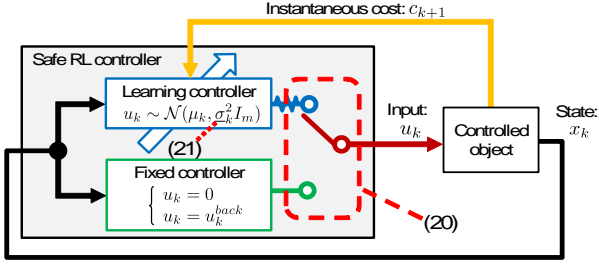}
	\vskip-2mm
	\caption{Block diagram of the closed loop system with our automatic exploration process adjustment method}
	\label{fig:block}
\end{figure}

In the rest of this paper, we assume that Assumptions~\ref{assumption1}-\ref{assumption4} hold. Based on the results described in Subsection~\ref{subsec_no_exp} and Lemma~\ref{cor6}, we propose the following control law as an automatic exploration process adjustment method for the Gaussian policy.
\mathindent=1em
\begin{align}
	\left\{\!\!
	\begin{array}{ll}
		u_k\sim \mathcal{N}(\mu_k,\underline{\sigma}^2_k I_{m}) & 
		\mathrm{if}\ x_k\in \mathcal{X} \land (\bar{x}_{k+1}^{\epsilon}\!\in\! \mathcal{X}^\mathrm{int},\forall \epsilon \in \mathcal{E})\\
		u_k=0 &\!\! \mathrm{else\ if} \ x_k\!\in\!\mathcal{X} \\
		u_k=u^{back}_k &\!\!  \mathrm{otherwise}
	\end{array}
	\right. \!\!\!\!,
	\label{eq:u_RL}
\end{align}
\mathindent=2em
where $\bar{x}_{k+1}^{\epsilon}:=Ax_k+B\mu_k+\epsilon$. In addition, we define $\underline{\sigma}_k$ in the top of (\ref{eq:u_RL}) as
\begin{align}
	\underline{\sigma}_k:=\min_{j, \epsilon}
	\frac{1}{\|h_{j}^\mathrm{T}B\|_2 \Phi^{-1}(\eta')} 
	\left( \dd_j\!-\!h_{j}^\mathrm{T}\bar{x}_{k+1}^{\epsilon}\right), \label{proposed_method}
\end{align}
where $\eta' := 1-\frac{1-\eta^\frac{1}{\tau}}{n_c}$. The block diagram of the closed loop system including  the safe RL controller based on our exploration process adjustment method is depicted in Fig.~\ref{fig:block}. As shown in this figure, the safe RL controller selects whether the exploratory input is used or not at each time according to the control law (\ref{eq:u_RL}) as well as adjusts the variance-covariance matrix of a normal distribution used in the Gaussian policy for exploration with $\underline{\sigma}_k$ in (\ref{proposed_method}). In other words, the safe RL controller contains two kinds of controllers; the learning controller with a time-varying variance-covariance matrix and the model-based fixed one, and selects either of them to generate the control input at each time\footnote{According to the terminology introduced in \cite{Okawa+19a}, this is classified as ``switching LeFiCo".}.

We have the following theorem regarding the satisfaction of the joint chance constraint with the above control law as an automatic exploration process adjustment method.
\setcounter{thm}{0}
\begin{thm} \label{theorem1}
	Let Assumptions~\ref{assumption1}-\ref{assumption4} hold and the input $u_k$ at time $k$ be selected according to~\eqref{eq:u_RL}. Then, the joint chance constraint~\eqref{eq:cc}, that is, $\CC{x_k}{\eta}$,  is satisfied at all time $k\ge 1$. 
\end{thm}
\begin{pf}\normalfont
	If $x_k \in \mathcal{X} \land (\bar{x}_{k+1}^{\epsilon}\!\in\! \mathcal{X}^\mathrm{int},\forall \epsilon \in \mathcal{E})$, $\CCp{x_{k+1}}{\eta'}$ is satisfied by selecting the inputs stochastically according to $u_k\sim\mathcal{N}(\mu_k,\underline{\sigma}^2_k I_{m})$ from Lemma~\ref{cor6}, and this is a sufficient condition for $\CC{x_{k+1}}{\eta^\frac{1}{\tau}}$. In addition, if $x_k\!\in\!\mathcal{X} \land \lnot(\bar{x}_{k+1}^{\epsilon}\!\in\! \mathcal{X}^\mathrm{int},\forall \epsilon \in \mathcal{E})$, $x_{k+1}$ always satisfies $x_{k+1}\in\mathcal{X}$ with the input $u_k=0$ from Assumption~\ref{assumption0}. Therefore, $\CC{x_{k+1}}{\eta^\frac{1}{\tau}}$ is satisfied at all time $k\geq 1$ with the control law~(\ref{eq:u_RL}) if $x_k\in \mathcal{X}$. Furthermore, this $\CC{x_{k+1}}{\eta^\frac{1}{\tau}}$ is a sufficient condition for $\CC{x_{k+1}}{\eta}$ since $\eta^{\frac{1}{\tau}}\geq\eta$ for any $\tau\in\mathbb{Z}_{\geq 1}$. 
	
	On the other hand, if $x_k\notin\mathcal{X}$, $k \geq 1$, Lemma~\ref{lem:tau} shows that $\CC{x_{k+1}}{\eta}$ is satisfied for an arbitrary $x_k \in \mathbb{R}^n$ with the input $u_k=u^{back}_k$ according (\ref{eq:u_RL}).
	
	Consequently, $\CC{x_k}{\eta}$ is satisfied at all time $k \ge 1$.\qed
\end{pf}

\begin{rem}\normalfont \label{rem1}
	As shown in the proof described in Appendix~\ref{proof_prop}, the pair (\ref{Eneq_Sigma}) and (\ref{Eneq_NonSigma})	is a sufficient condition for the state $x_{k+1}$ at the next time $k+1$ to satisfy the joint chance constraint (\ref{eq:cc}) for more general $\Sigma$. In this case, however, it is difficult to derive $\Sigma$ which satisfies this condition.
\end{rem}

\begin{rem}\normalfont \label{rem2}
	Theorem~\ref{theorem1} requires that the controlled object satisfies Assumption~\ref{assumption0}. 	However, there 	may not be many systems satisfying this assumption by themselves. For this problem, we can expand the applicability of our proposed method by forming a minor feedback loop with some kind of pre-designed controllers based on known information of the controlled object. We can say that this is one of the advantages of the control approach to use RL and model-based control simultaneously in parallel (``parallel LeFiCo") as discussed in \cite{Okawa+19a}.
\end{rem}

\subsection{Safe learning algorithm with the automatic exploration process adjustment}\label{SafeRLAlg}
\begin{algorithm}[t]
	\caption{Safe one-step actor-critic algorithm with automatic exploration process adjustment} \label{alg1}
	\begin{algorithmic}[1]
		\Initialize{ $\theta \gets \theta_0$, $\ \ww \gets \ww_0$, \ $\iota \gets 1$}
		\Loop \ (Execute below at $t=0$ and every $T_s$ time)
		\State Observe $x$
		\If{$t\neq 0$}
		\State Get cost $c$
		\State $\delta\gets -c + \gamma \hat{V}(x; \theta) - \hat{V}(x^{-}; \theta)$
		\State $\theta \gets \theta +\alpha \delta \frac{\partial \hat{V}}{\partial \theta}(x^{-} ;	\theta)$
		\State $\ww \gets \ww \!+\! \beta \iota \delta \frac{\partial \log \Pi}{\partial \ww}(u^{-}|x^{-} ; \ww) $
		\EndIf
		\State Select $u$ according to (\ref{eq:u_RL}) \label{alg:lno:setu}
		\State Input $u$ to controlled object
		\State $x^{-} \gets x$,\ $u^{-} \gets u$, \ $\iota \gets \gamma \iota$
		\EndLoop
	\end{algorithmic}
\end{algorithm}
Algorithm~\ref{alg1} shows a concrete example of the proposed safe RL method, which is the algorithm of the one-step actor-critic described in the Section 13 of \cite{Sutton+18a} with our automatic exploration process adjustment method shown in the previous subsection. In this algorithm, $\hat{V}(x; \theta)$ is the estimated value of the state value function, $\theta\in\mathbb{R}^{N_{\theta}}$ is the sate value weight, and $N_{\theta}$ is its number. In addition, $\delta\in\mathbb{R}$ is the TD (temporal difference) error, and $\alpha\in[0,1)$ and $\beta\in[0,1)$ are the learning rates (step sizes). In this algorithm, the inputs are selected according to the control law~(\ref{eq:u_RL}) to adjust its exploration process automatically at each time as shown in Line \ref{alg:lno:setu}, and then the state value weight $\theta$ and the policy parameter $\ww$ are updated. As a result, we achieve to learn the appropriate policy by updating policy parameters 
with the satisfaction of the joint chance constraint (\ref{eq:cc}) for the pre-specified $\eta$ guaranteeing that the probability of constraint satisfaction is equal to or greater than its pre-specified lower bound.

\subsection{Comparison with related work}\label{Rela}
In the field of RL, the concept of ``safety" (or its opponent, ``risk") is defined according to many kinds of formulation. Here we restrict our discussions to the ones regarding continuous state and action spaces as ours. For example, \cite{Wen+18a} proposed an RL method in which constraints satisfied for safety are defined as the expected cost over finite-length trajectories. In addition, some RL methods have been proposed to guarantee their safety even during learning. \cite{Berkenkamp+17a} used a known policy which guarantees safety to prevent the system from getting into an unrecoverable or undesired situation. Also, \cite{Achiam+17a} proposed a policy search algorithm for an RL problem in a constrained Markov Decision Process, 
which guarantees constraint satisfaction throughout training. However, these studies do not deal with satisfaction of the constraints explicitly defined in their control problems as their safety.

The following three methods regarding safe RL are the most similar to our study in the sense that they guarantee their safety during learning from viewpoint of satisfying the constraints explicitly defined in their control problems. \cite{Dalal+18a} showed a safe exploration method for RL algorithms to satisfy its safety constraints if one-time initial pre-training of a model can be used. \cite{Li+18a} proposed a safe RL framework with a supervisory element between the RL agent and the linear control system. Furthermore, \cite {Cheng+19a} showed how to modify existing RL algorithms to guarantee safety of the nonlinear system whose dynamics consists of partially known autonomous dynamics and completely known actuated one.
However, as compared with these existing methods, the safe RL method with our exploration process adjustment method guarantees its safety theoretically even if we can only use partial information of both autonomous and actuated dynamics of the nonlinear system: a linear approximate (nominal) model and upper bounds of its approximation error.

\section{Simulation Verification}\label{Simu}
This section verifies the validity and effectiveness of the safe RL method with our automatic exploration process adjustment method through numerical simulation.

\subsection{Simulation condition}\label{subsec:simcon}
\subsubsection{Control objective and constraints}
Let us consider the following nonlinear function $f:\mathbb{R}^2 \rightarrow \mathbb{R}^2$:
\begin{align}
	f(x)
	\!=\!
	\left[\!
	\begin{array}{c}
		\!f_{1}(x)\! \\
		\!f_{2}(x)\!
	\end{array}
	\!\right]
	\!=\!
	\left[\!
	\begin{array}{c}
		0.3 x_1 - 0.4 \sin x_2 \\
		-0.1 x_2 + 0.2 \cos x_1 \!-\! 0.2
	\end{array}
	\!\right], 
	\label{eq:sim_f}
\end{align}
where $x=[x_1,x_2]^\mathrm{T}\in\mathbb{R}^2$ and $f(0)=0$. Regarding this function, the Jacobian matrix is given by
\begin{align*}
	\frac{\partial f}{\partial x}(x)=
	\left[
	\begin{array}{cc}
		0.3 & -0.4 \cos x_2\\
		-0.2\sin x_1 & -0.1
	\end{array}
	\right].
\end{align*}
Therefore, the Frobenius norm $\|\frac{\partial f}{\partial x}(x)\|_F$ of $\frac{\partial f}{\partial x}(x)$ satisfies
\begin{align*}
	&\left \|\frac{\partial f}{\partial x}(x) \right\|_F
	=\sqrt{0.04 \sin^2 x_1 + 0.16 \cos^2 x_2 + 0.1}<1
\end{align*}
for an arbitrary $x\in\mathbb{R}^2$ since $|\sin x_1|\leq1$ and $|\cos x_2|\leq1$. This means that $f$ is a contraction mapping in entire $\mathbb{R}^{2}$ with the origin being its fixed point.

We let $x_k=[x_{1_k},x_{2_k}]^\mathrm{T}\in\mathbb{R}^2$ and $u_k\in\mathbb{R}$ be the state and the input at time $k$, respectively, and consider the following discrete-time affine nonlinear system: 
\begin{align}
	x_{k+1}=f(x_k)+g u_k, 
	\label{eq:sim_true}
\end{align}
where $f$ is the function described above and $g=[1,1]^\mathrm{T}$. In addition, we use the following linear nominal model: 
\begin{align}
	x_{k+1} \simeq Ax_k\!+\!bu_k,\ 
	A\!=\!
	\left[\!\begin{array}{cc}
		0.3 & 0 \\
		0 & -0.1
	\end{array}\!\right],\ 
	b\!=\!
	\left[\!\begin{array}{c}
		1 \\
		1
	\end{array}\!\right].
	\label{eq:sim_nom_AB}
\end{align}
According to Assumption~\ref{assumption1}, we assume $A$ and $b$ are known while $f$ and $g$ are unknown. The upper bounds of the error function $e_i$ given in (\ref{eq:err_fun}) become
\begin{align*}
	\hspace{-1.6em}
	\sup_{x\in\mathbb{R}^2,u\in\mathbb{R}}\hspace{-1em}|-0.4 \sin x_2| =0.4, 
	\hspace{-0.1em}
	\sup_{x\in\mathbb{R}^2,u\in\mathbb{R}}\hspace{-1em}|0.2 \cos x_1-0.2| =0.4. 
\end{align*}
Therefore, we let $\bar{e}_1=0.4$ and $\bar{e}_2=0.4$, and assume that these values are previously known to satisfy Assumption~\ref{assumption2}.

On the other hand, we let $|x_{1}|\leq 10$ be the constraints. That is, the set $\mathcal{X}$ is given by $\mathcal{X}=\left\{x\in\mathbb{R}^2 \,|\, H x \preceq \dd \right\}$, where
\begin{align*}
	H\!=\!
	\left[\!\begin{array}{c}
		h_1^\mathrm{T} \\
		h_2^\mathrm{T}
	\end{array}\!
	\right]
	\!=\!
	\left[\!\begin{array}{cc}
		1 & 0\\
		-1 & 0
	\end{array}\!
	\right],\ 
	\dd
	\!=\!
	\left[\!\begin{array}{c}
		\dd_1 \\
		\dd_2
	\end{array}
	\!\right]
	\!=\!
	\left[\!\begin{array}{cc}
		10\\
		10
	\end{array}
	\!\right]. 
\end{align*}
Assumption~\ref{assumption0} is satisfied since $f$ is a contraction mapping in $\mathbb{R}^{2}$ with the origin being its fixed point as described above  and $\mathcal{X}$ is a convex region containing the origin.  We also let the initial state be $x_0=[5,5]^\mathrm{T}\in\mathcal{X}$.

Since $h^\mathrm{T}_j b\neq 0$, $j\in\{1,2\}$, the pair of the above nominal model and the constraints satisfies Assumption~\ref{assumption4}.

In addition, we assume that the instantaneous cost $c_{k+1}=\ell(x_k,u_k)$ can be measured directly at each time, while the following cost function $\ell(\cdot,\cdot)$ is unknown:
\mathindent=1em
\begin{align}
	\ell(x_k,u_k)
	&=x_{k+1}^\mathrm{T}Q x_{k+1} +R u_k^2\cr
	&=(f(x_k)+g u_k)^\mathrm{T}Q(f(x_k)+g u_k) +R u_k^2,
	\label{eq:sim_inst_cost}
\end{align}
\mathindent=2em
where $Q=1.0\times 10^5 I_2$ and $R=1$.

\subsubsection{Control input determination with the proposed exploration process adjustment method}
In this verification, we update a policy according to Algorithm~\ref{alg1} described in Subsection \ref{SafeRLAlg}. Specifically, we define $T(=15)$ steps as $1$ episode and learn the policy to determine inputs which minimize the cumulative cost $J=\sum_{k=0}^{T-1}c_{k+1}$ of the instantaneous cost given in (\ref{eq:sim_inst_cost}) in each episode starting from the initial state $x_0$.

The estimated state value function $\hat{V}(x;\theta)$ and the mean $\mu(x;\ww)$ of the input $u$ are, respectively, given by 
\begin{align}
	\hat{V}(x; \theta)=\sum_{i=1}^{N_\theta} \phi_i(x)\theta_i,\ 
	\mu(x;\ww) =\sum_{i=1}^{N_\ww} \phi_i(x) \ww_{i},
	\label{eq:sim_mu}
\end{align}
where $\theta = [\theta_{1}, \dots, \theta_{N_\theta}]^\mathrm{T} \in \mathbb{R}^{N_\theta}$ and $\ww = \left[\ww_{1}, \dots, \ww_{N_{\ww}}\right]^\mathrm{T} \in \mathbb{R}^{N_{\ww}}$. In the above equation, $\phi_i: \mathbb{R}^{2} \to \mathbb{R}$ represent the feature extractors given by the following Gaussian radial basis functions:
\begin{align}
	\phi_i(x)=\exp\left(-\frac{\|x-\varrho_i\|^2}{2\varsigma_i^2}\right),\ i=1, \ldots, N_\theta, 
	\label{basis}
\end{align}
where $\varrho_i \in \mathbb{R}^{2}$ and $\varsigma_i^2 > 0$ are the central points and the variances of each basis function, respectively. We calculate the mean value $\mu_k=\mu(x_k;\ww_k)$ according to (\ref{eq:sim_mu}) with the state $x_k$ at each time and the policy parameters $\ww_k$, and then, select the input $u_k$ according to the control law~(\ref{eq:u_RL}). On the other hand, the pair of the coefficient matrix $(A,b)$ of the nominal model in (\ref{eq:sim_nom_AB}) is controllable and its controllability index is $2$. From this result, we calculate the inputs $u_k^{back}$ in the control law~(\ref{eq:u_RL})  as follows:
\begin{align}
	\left[
	u_{k}^{back},
	u_{k+1}^{back}
	\right]^{\mathrm{T}}
	\!=
	\tilde{B}^{-1}(x_0 \!-\!A^2 x_k),\ 
	\tilde{B}:=[Ab, b].
\end{align}
These inputs are designed so as to make the state go back to the initial point. Indeed, these two successive inputs make any state $x \notin \mathcal{X}$ return into $\mathcal{X}$ within $2$ steps. Therefore, Assumption \ref{assum_tau} is satisfied. The state value weight $\theta$ and the policy parameter $\ww$ are updated with the instantaneous cost given by (\ref{eq:sim_inst_cost}) at each time. The values of each parameter are listed in Table~\ref{tb:1}.

\begin{table}[t!]
	\centering
	\caption{Simulation parameters \label{tb:1}}
	\scalebox{0.95}{
		\begin{tabular}{ccc}
			\hline
			Symbol  & Definition  & Value  \\
			\hline \hline
			$T$ & Number of simulation steps & $15$\\
			$\gamma$ & Discount rate  & $1.0$\\
			$\alpha,\ \beta$ & Learning rates & $1.0\times 10^{-10}$\\
			$N_\theta, N_\ww$ & Number of learning parameters& $121$\\
			$N$ & Number of learning episodes & $1.5\times 10^4$\\
			$\eta$ & $\begin{array}{c}\mathrm{Lower\ bound\ of\ probability}\\ \mathrm{of\ constraint\ satisfaction}\end{array}$ & $0.95$\\
			\hline
		\end{tabular}
	}
\end{table}

\subsection{Simulation result}
\begin{figure}[t]
	\centering
	\includegraphics[clip,width=6.5cm]{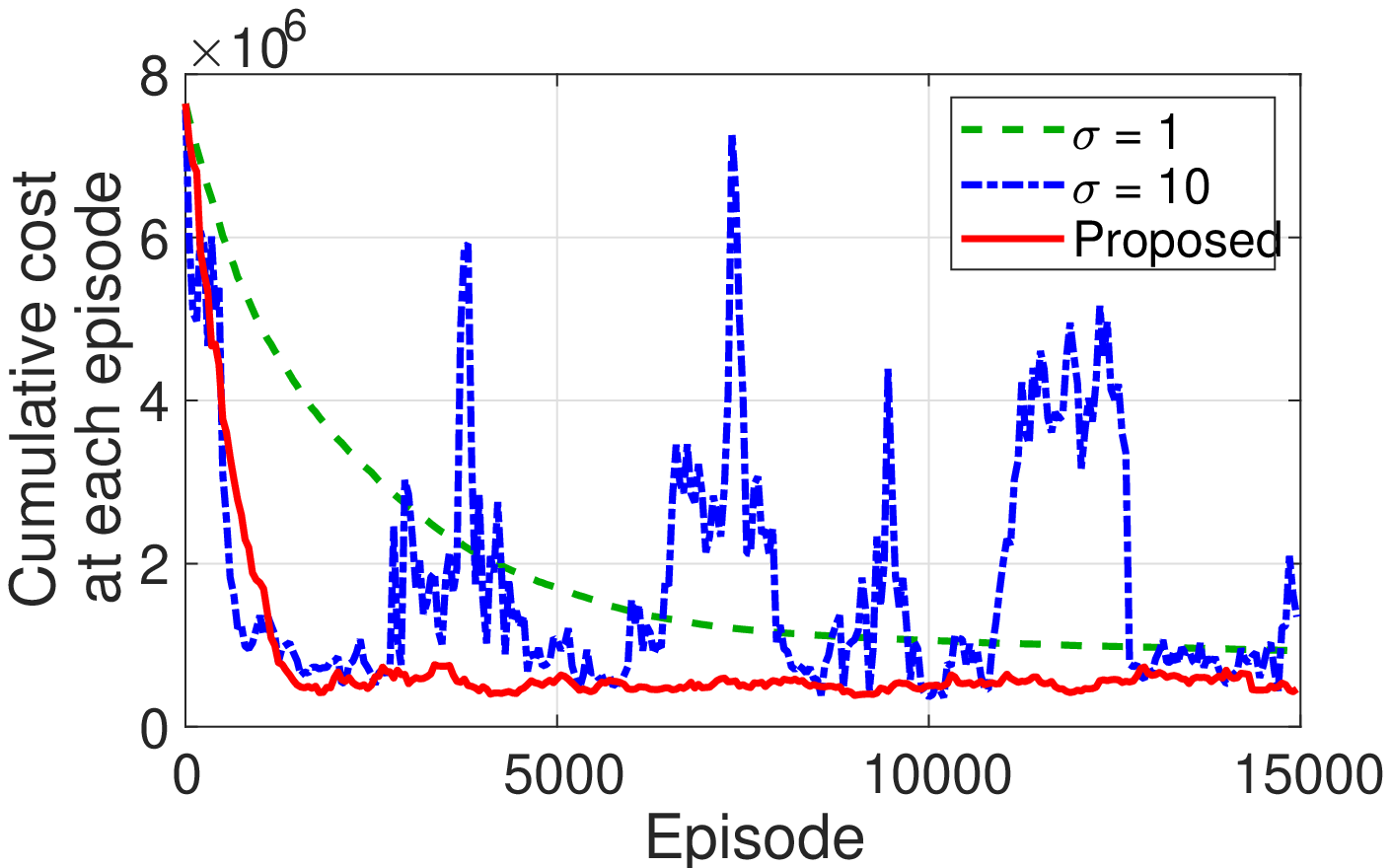}
	\vskip -2mm
	\caption{Cumulative cost at each episode}
	\label{fig:cost}
	\vspace{2pt}
	\includegraphics[clip,width=6.5cm]{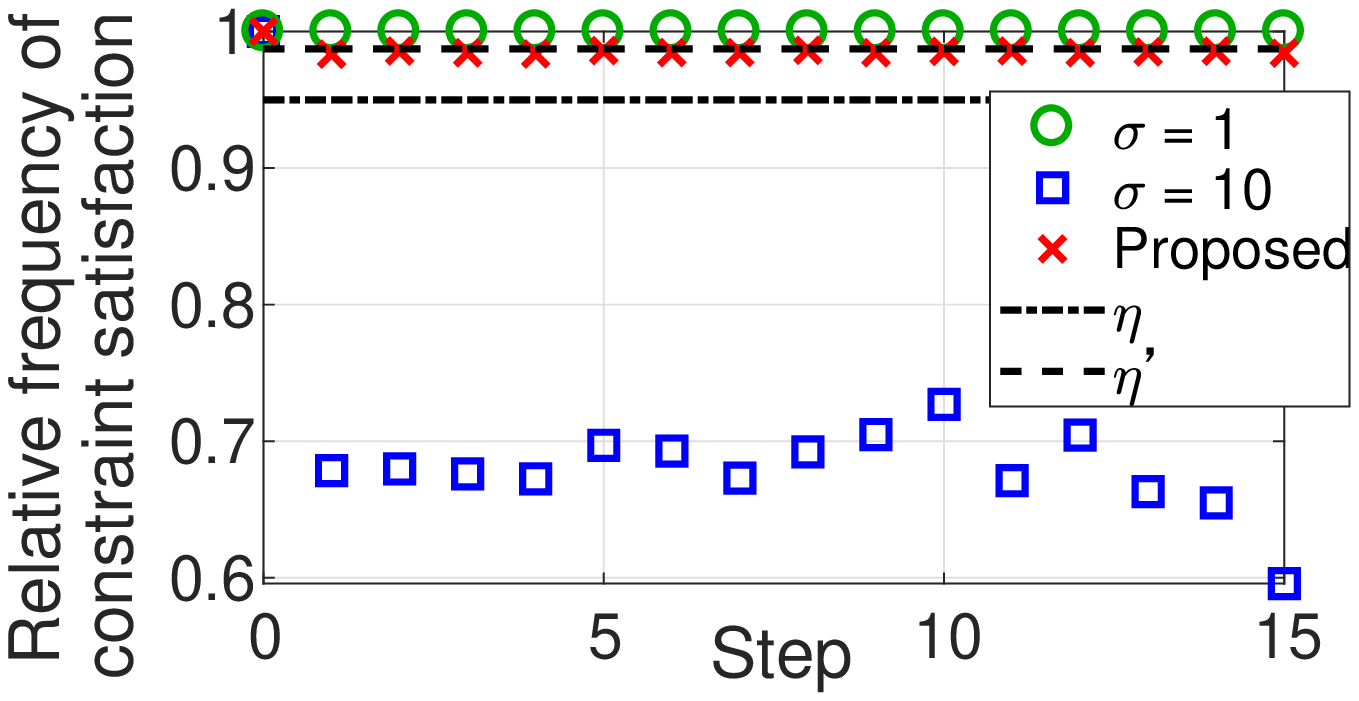}
	\vskip -2mm
	\caption{Relative frequencies of constraint satisfaction}
	\label{fig:eta}
	\vspace{2pt}
	\includegraphics[clip,width=6.5cm]{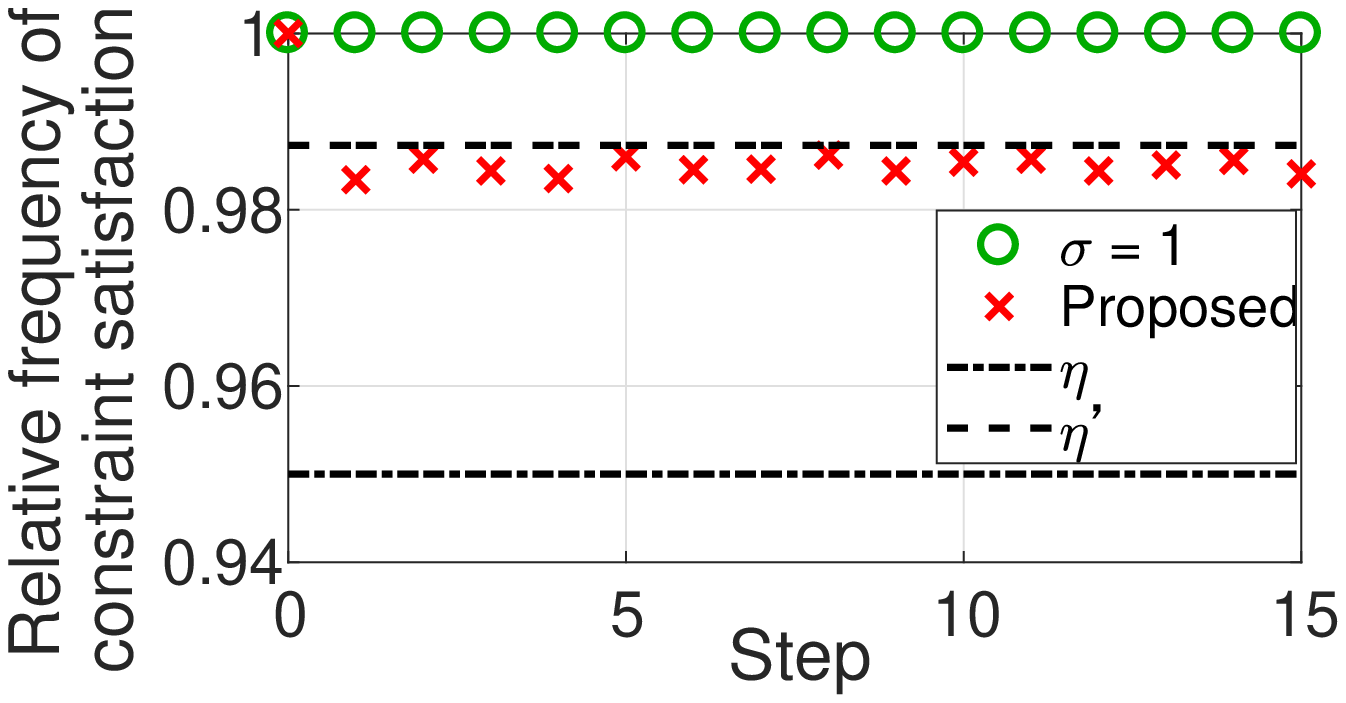}
	\vskip -2mm
	\caption{Enlarged figure of Fig.~\ref{fig:eta}}
	\label{fig:eta_en}
\end{figure}

\begin{figure}[t]
	\centering
	\includegraphics[clip,width=6.5cm,height=3.5cm]{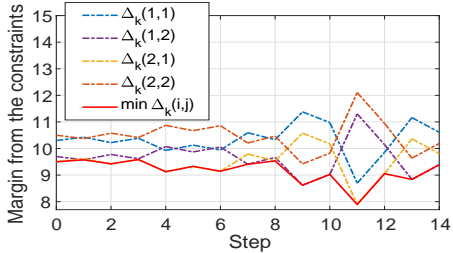}
	\caption{Results of margins from the constraints and its minimum at the final episode}
	\label{fig:dist}
\end{figure}

\begin{figure}[t]
	\centering
	\includegraphics[clip,width=6cm]{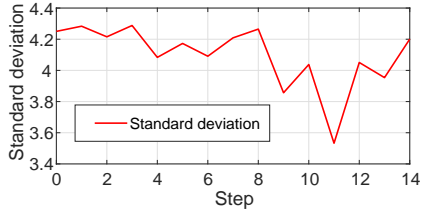}
	\caption{Result of $\underline{\sigma}_k$ at the final episode}
	\label{fig:sig}
\end{figure}

\begin{figure}[t]
	\centering
	\includegraphics[clip,width=6.5cm,height=3.5cm]{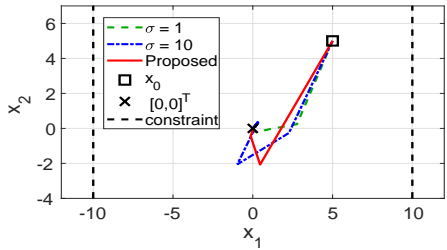}
	\caption{Result of the trajectory}
	\label{fig:tra}
\end{figure}

The results of this verification are shown in Figs.~\ref{fig:cost}-\ref{fig:tra}. Fig.~\ref{fig:cost} shows the cumulative cost $J$ obtained without exploration, that is, letting the inputs $u_k$ at each step time be their mean values $\mu_k$ calculated with the policy parameters $w_k$, at every $50$ episodes. Fig.~\ref{fig:eta} shows the relative frequencies of the constraint satisfaction with respect to each time which are obtained by dividing the number of the episodes satisfying the constraints by its total number $N$,  and Fig.~\ref{fig:eta_en} is the enlarged figure of Fig.~\ref{fig:eta}. Since the total number of episodes $N$ is large enough and the occurrence of constraint satisfaction between arbitrary two episodes is independent of each other, we use relative frequencies to confirm the validity of Theorem~\ref{theorem1}. In addition, Fig.~\ref{fig:dist}  shows the margins from the constraints and its minimum value at the final episode. Due to the simulation conditions described in the previous subsection, the margins $\dd_j-h^\mathrm{T}_j(Ax_k+B\mu_k +\epsilon)$, $\epsilon\in\mathcal{E}$, $j\in\{1,2\}$ can be summarized into four scalar values. We denote them by $\Delta_k(i,j)$ where $i=1$ if $\epsilon=[\bar{e}_1,\ \pm\bar{e}_2]^{\mathrm{T}}$ and $i=2$ if $\epsilon=[-\bar{e}_1,\ \pm\bar{e}_2]^{\mathrm{T}}$, and show $\Delta_k(i,j)$ in dashed lines and its minimum value at each step time in a solid line in this figure, respectively. Fig.~\ref{fig:sig} shows $\underline{\sigma}_k$ given in (\ref{proposed_method}) at the final episode. Furthermore, Fig.~\ref{fig:tra} shows the trajectory of the state with the policy parameter at the final episode.  Among these figures, in Figs.~\ref{fig:cost}-\ref{fig:eta_en} and \ref{fig:tra}, the results by using the Gaussian policy with the fixed standard deviation $\sigma=1$ are shown in green, those with the bigger fixed standard deviation $\sigma=10$ are in blue, and those by using our proposed method are in red. Furthermore, in this verification, if $|x_{1_k}|>10$, $k=1,2,\ldots,T$ at each episode, we gave a penalty $(T-k+1)c_{k+1}$ and let the estimated state value be $\hat{V}(x)=0$ to terminate the current episode, and then, start another episode from the initial state when we use the Gaussian policy with fixed standard deviations.

We can confirm that the relative frequencies of the constraint satisfaction with the fixed standard deviation $\sigma=1$ in green is greater than the lower bound of the probability of constraint satisfaction $\eta = 0.95$ in Fig.~\ref{fig:eta}; however, as shown in Fig.~\ref{fig:cost}, the corresponding result of the cumulative cost decreases slowly as the number of episodes increases. On the other hand, the results with the fixed standard deviation $\sigma=10$ in blue show that, though the cumulative cost decreases efficiently in Fig.~\ref{fig:cost}, there are some step times when the relative frequency of the constraint satisfaction becomes lower than $\eta$. As compared with these results, our proposed method achieves not only to decease its cumulative cost efficiently but also to guarantee that the relative frequencies of the constraint satisfaction are greater than $\eta=0.95$ at all step times, while there are some step times when its value become lower than $\eta'=1-\frac{1-\eta^{\frac{1}{2}}}{2} \approx 0.987$ which is used to derive the standard deviation $\underline{\sigma}_k$ in (\ref{proposed_method}) as shown in Fig.~\ref{fig:eta_en}. 

In addition, it is shown that the standard deviation $\underline{\sigma}_k$ in Fig.~\ref{fig:sig} corresponds to the minimum value among $\Delta_k(i,j)$ at each step time in a solid line in Fig.~\ref{fig:dist}. We can confirm from this result that our proposed method automatically adjusts its exploration process by changing the variance $\underline{\sigma}_k^2$ in the Gaussian policy at each step time, even if the input is used for exploration.

Furthermore, from the control result with the policy parameters at the final episode shown in Fig.~\ref{fig:tra}, the learned policy parameter enables us to transit the state $x_k$ of the system into the origin $x=[0,0]^\mathrm{T}$, which is the fixed point of $f(x)$ and the instantaneous cost given in  (\ref{eq:sim_inst_cost}) becomes $0$ with the input $u_k=0$.


\section{Conclusion and Future Work}\label{Conc}

In this paper, we considered an RL problem in continuous state and action spaces with constraints explicitly defined in the control problem, and  proposed an automatic exploration process adjustment method for safe RL which achieves the satisfaction of a joint chance constraint derived from the above-mentioned constraints. More specifically, our method adjusts the exploration process automatically utilizing a known linear nominal model of the controlled object,  and we theoretically showed that this method attains the pre-specified lower bound of the constraint satisfaction probability. We also verified the validity and effectiveness of our method through numerical simulations and showed that it achieves to learn its policy appropriately by guaranteeing the satisfaction of the joint chance constraint with the pre-specified probability.

One of the future work is to develop a safe RL method for control problems in which reinforcement learning methods and model-based control methods are used simultaneously with different sampling times. We have to consider this situation since updating huge number of parameters in an RL method requires more computation in general, and thus, it is required to take a longer time to determine its control input. However, in such a situation,  the probability of constraint satisfaction is not guaranteed with the exploration process adjustment method presented in this paper since the state transition with the control inputs from a model-based control method between the sampling times of an RL method is not considered. Therefore, we plan to develop an automatic exploration process adjustment method to guarantee the satisfaction of the constraints even in such a situation.

\bibliography{rl}

\appendix
\section{Property of Markov chain} \label{sec:proof_tau} 
Let $\tau$ be a natural number greater than or equal to $1$. Also, let a stochastic process $\{ X_k \}$ be a discrete-time Markov chain whose state space and transition probability matrix, respectively, are  $\{1, 2, \ldots, \tau+1\}$ and 
\[ \left[ \begin{array}{ccccc}
	\rho_1    & 1-\rho_1 & 0        & \cdots & 0\\
	\rho_2    &    0     & 1-\rho_2 & \cdots & 0 \\
	\vdots    & \vdots   & \vdots   & \ddots & \vdots\\
	\rho_\tau &    0     &   0      & \cdots       & 1-\rho_\tau\\
	1         &    0     &   0      & \cdots& 0  \\
\end{array} \right]. \]
\setcounter{thm}{0}
\begin{prop} \label{prop:tau} \normalfont
	Consider the above Markov chain $\{ X_k \}$. If $\mathrm{Pr}\left\{ X_0 = 1 \right\} = 1$, then $\mathrm{Pr}\left\{ X_k = 1 \right\} \geq \rho_1^\tau$ for all $k \in \mathbb{Z}_{\geq 0}$.
\end{prop}
\begin{pf}
	We denote by $p_k^{(i)} := \mathrm{Pr}\left\{ X_k = i \right\}$ the probability that the state is $i$ at time $k$ and prove $p_k^{(1)} \geq \rho_1^\tau$ by induction. Firstly, we consider the case in which the state transits within State $1$. 
	In this case, $p_t^{(1)} \geq p_0^{(1)}\rho_1^t \geq \rho_1^\tau$ ($t=0,1,\ldots, \tau$) holds since $p_0^{(1)} = 1$.
	
	Next, we assume all values less than $k+1 (> \tau)$ is true. We obtain the following simultaneous recurrence formulas:
	\begin{align*} \left\{ \begin{array}{rl}
			p_{k+1}^{(1)} &= p_k^{(\tau+1)} + \sum_{i=1}^{\tau} \rho_i p_k^{(i)}
			= \sum_{i=1}^{\tau+1} \rho_i p_k^{(i)}\\
			p_{k+1}^{(i)} &= (1-\rho_{i-1}) p_k^{(i-1)} \qquad (i=2,3, \ldots, \tau+1) \\
		\end{array} \right. , \end{align*}
	where $\rho_{\tau+1} := 1$. Now,
	\begin{align*}
		p_{k}^{(i)}
		&= (1-\rho_{i-1}) p_{k-1}^{(i-1)}
		= (1-\rho_{i-1}) (1-\rho_{i-2}) p_{k-2}^{(i-2)} \\
		&= \cdots
		= \Big\{\prod_{j=1}^{i-1} (1 -\rho_j) \Big\} p_{k-i+1}^{(1)},
	\end{align*}
	and thus, we obtain
	\begin{align*}
		p_{k+1}^{(1)} &=
		\sum_{i=1}^{\tau+1} \Big\{\rho_i \prod_{j=1}^{i-1} (1 -\rho_j) \Big\} p_{k-i+1}^{(1)}.
	\end{align*}
	Since the sum of the coefficient in the right-hand side of the above equation is $1$, the assertion of the proposition holds at $k+1$ from the assumption of induction.\qed
\end{pf}

\section{Property of Gaussian policy} \label{proof_prop}

\begin{prop}\normalfont\label{prop}
	Let Assumptions \ref{assumption1} and  \ref{assumption2} hold.	For the state $x_k \in \XX$ at time $k$ of the nonlinear system given in (\ref{eq:true_model}), we select the input $u_k \in \mathbb{R}^{m}$ according to the $m$-dimensional normal distributed function with the mean $\mu_k \in \mathbb{R}^{m}$ and the covariance $\Sigma \in \mathbb{R}^{m \times m}$ given in (\ref{eq:Sigma}). Then, the inequality condition
	\begin{align}
		\sigma
		\leq
		\frac{1}{\|h_{j}^\mathrm{T}B \|_2 \Phi^{-1}({\etaspade}) } \left \{  \dd_j\!-\!h_{j}^\mathrm{T}(Ax_{k}\!+\!B\mu_k\!+\!\epsilon) \right \}, \nonumber\\
		\forall j \ \mathrm{ s.t. } \ h_{j}^\mathrm{T}B \neq 0, \ \forall \epsilon \in \mathcal{E} \label{Eneq_sigma}, \\
		h_{j}^\mathrm{T}(Ax_{k} + \epsilon)  
		\leq
		\dd_j, \forall j \ \mathrm{ s.t. } \ h_{j}^\mathrm{T}B = 0, \ \forall \epsilon \in \mathcal{E}
	\end{align}
	is the sufficient condition to satisfy $\CCp{x_{k+1}}{\etaspade}$, where $\Phi(\cdot)$ is the cumulative distribution function of a zero mean unit variance Gaussian random variable, and $\etaspade \in (0.5, 1)$.
\end{prop}
\begin{pf}
	Let us denote the nominal value $\hat{x}_{k+1}$ at time $k+1$ of the nonlinear system by
	\begin{align}
		\hat{x}_{k+1} := A x_k +Bu_k. 
	\end{align}
	With this value, we obtain
	\begin{align}
		\hat{x}_{k+1} \sim \mathcal{N}(Ax_k+B\mu_k, B^\mathrm{T}\Sigma B).
	\end{align}
	Firstly, let us select an arbitrary $j$. In this case, 
	\begin{align*}
		\CCp{x_{k+1}}{\etaspade} \nonumber 
		&\Leftrightarrow \ \mathrm{Pr}\{h_{j}^\mathrm{T}x_{k+1} \leq \dd_{j}\} \geq \etaspade \nonumber \\
		&\Leftarrow
		\ \mathrm{Pr}\{h_{j}^\mathrm{T}(\hat{x}_{k+1}+\epsilon) \leq \dd_{j}\} \geq \etaspade,\ \forall \epsilon \in \mathcal{E}. 
	\end{align*}
	Next, let us select an arbitrary $\epsilon \in \mathcal{E}$. In this case, the above chance constraint  becomes the following deterministic constraint \citep{Boyd04+a}: 
	\mathindent=1em
	\begin{align}
		&\ \mathrm{Pr}\{h_{j}^\mathrm{T}(\hat{x}_{k+1}+\epsilon) \leq \dd_{j}\} \geq {\etaspade} \nonumber \\
		&\Leftrightarrow 
		\dd_{j}\!-\!h_{j}^\mathrm{T}(Ax_{k}\!+\!B\mu_k\!+\!\epsilon) \!\geq\! \Phi^{-1}({\etaspade}) \left\|h_{j}^\mathrm{T}B \Sigma^{\frac{1}{2}}\right\|_2.  
		\label{eq:sig_con_ori}
	\end{align}
	\mathindent=2em
	Therefore, the inequality condition 
	\begin{align}
		\Phi^{-1}({\etaspade}) \|h_{j}^\mathrm{T}B \Sigma^{\frac{1}{2}} \|_2  \leq 
		\dd_j\!-\!h_{j}^\mathrm{T}(Ax_{k}\!+\!B\mu_k\!+\!\epsilon),  \nonumber\\
		\forall j=1,2,\ldots,n_c,\ \forall \epsilon \in \mathcal{E}
		\label{eq:sig_con}
	\end{align}
	is a sufficient condition for the state $x_{k+1}$ to satisfy $\CCp{x_{k+1}}{\etaspade}$ at time $k+1$. 
	By dividing the above inequality condition depending on whether $h_{j}^\mathrm{T}B = 0$ or not, we obtain with $0 < \Phi^{-1}({\etaspade}) < \infty$ that
	\begin{align}
		\|h_{j}^\mathrm{T}B \Sigma^{\frac{1}{2}} \|_2  
		\leq 
		\frac{1}{\Phi^{-1}({\etaspade}) }\! \left \{  \dd_j\!-\!h_{j}^\mathrm{T}(Ax_{k}\!+\!B\mu_k\!+\!\epsilon) \right \} \nonumber\\
		\forall j \ \mathrm{ s.t. } \ h_{j}^\mathrm{T}B \neq 0, \ \forall \epsilon \in \mathcal{E}, \label{Eneq_Sigma} \\
		h_{j}^\mathrm{T}(Ax_{k} + \epsilon)  
		\leq 
		\dd_j,  \forall j \ \mathrm{ s.t. } \ h_{j}^\mathrm{T}B = 0, \ \forall \epsilon \in \mathcal{E}. \label{Eneq_NonSigma}
	\end{align}
	In addition, with the variance-covariance matrix given in (\ref{eq:Sigma}), we obtain
	\begin{align}
		\left\|h_{j}^\mathrm{T}B \Sigma ^{\frac{1}{2}}\right\|_2 = \sigma \|h_{j}^\mathrm{T}B\|_2.
	\end{align}
	Therefore, by substituting this into (\ref{Eneq_Sigma}) with respect to $j$ which satisfies $h_{j}^\mathrm{T} B \neq 0$ and dividing its both side by $\|h_{j}^\mathrm{T}B\|_2$, we obtain the condition in (\ref{Eneq_sigma}).\qed
\end{pf}

\begin{cor}\label{cor5} \normalfont
	Let assumptions used in Proposition~\ref{prop} hold. In addition, Assumption \ref{assumption4} holds. Then,
	\begin{align}
		\sigma \leq 
		\frac{1}{\|h_{j}^\mathrm{T}B\|_2 \Phi^{-1}(\etaspade)}\left \{ \dd_j\!-\!h_{j}^\mathrm{T}(Ax_{k}\!+\!B\mu_k\!+\!\epsilon)  \right \} \nonumber\\
		\forall j=1,2,\ldots,n_c,\ \forall \epsilon \in \mathcal{E}
	\end{align}
	is a sufficient condition to satisfy $\CCp{x_{k+1}}{\etaspade}$.
\end{cor}

\begin{pf}
	We can prove this corollary from  Proposition~\ref{prop} and the assumption $h_{j}^\mathrm{T} B \neq 0, \forall j = 1,2,\dots, n_{c}$.\qed
\end{pf}

\end{document}